\documentclass[cameraready]{Interspeech}

\title{Earnings25: A Comprehensive 500-Hour Speech Benchmark for Finance}

\author{Denglin}{Jiang}
\author{Haoran}{Zhou}
\author{Anshul}{Wadhawan}
\author{Brendan}{Fahy}
\author{Vinay}{Ramesh}
\author{David}{Weisberg}
\author{Dmitriy}{Derkachevskiy}
\author{Helen}{Sheehan}
\author{Srivas}{Prasad}
\author{Michele}{Franceschini}

\address{Bloomberg, United States}

\email{%
\{djiang108,hzhou245,awadhawan9,bfahy2,vramesh7,dweisberg6,
dderkachevsk,hsheehan,sprasad60,mfrancesch10\}@bloomberg.net%
}

\begin{document}
\maketitle

\begin{abstract}
We introduce Earnings25, a finance-domain benchmark for
evaluating automatic speech recognition (ASR) on English-language
earnings calls under realistic conditions. Earnings25 comprises two
complementary test sets: (i) \textbf{testset-full}, 498 hours of full
English-language S\&P 500 earnings calls from Q4 2025, and
(ii) \textbf{testset-segmented}, a 46-hour industry-balanced set of 290
segments sampled from English-language U.S. earnings calls in 2025.
The benchmark provides aligned transcripts and structured metadata,
including speaker roles, industry labels, and call structure, enabling
speaker- and industry-aware evaluation beyond aggregate word error rate
(WER). We report reproducible baselines for Whisper and Parakeet-TDT
using standardized scoring.
\end{abstract}

\section{Introduction}

\subsection{Financial speech understanding is challenging}
Earnings calls are a critical channel for corporate communication and pose unique challenges for automatic speech recognition (ASR). They combine spontaneous dialogue with scripted remarks, heavy financial jargon, company and product names, frequent numeric expressions, rapid turn-taking, and overlapping speakers during Q\&A—conditions that expose limitations of ASR systems trained primarily on read or general conversational speech.

These challenges are amplified by domain shift. Earnings calls often include operator boilerplate, variable recording quality due to telephony compression or background noise, and frequent speaker transitions. Linguistically, the domain contains jargon and acronyms under-represented in general ASR training data, as well as accented English from international executives. Although self-supervised and weakly supervised pretraining has improved general ASR performance (e.g., wav2vec~2.0~\cite{baevski2020wav2vec2} and Whisper~\cite{radford2022whisper}), domain mismatch remains a major source of error, motivating the need for recent, domain-specific speech resources.

\subsection{Toward a comprehensive benchmark for financial ASR}
Despite growing interest in financial speech and NLP, few publicly available resources have been purpose-built for \emph{evaluating} ASR on earnings calls. Existing datasets often emphasize training data without standardized evaluation protocols or lack the metadata required for detailed error analysis. As a result, the field lacks a principled and reproducible benchmark for measuring progress on financial ASR.

Beyond transcription accuracy, practical financial speech systems require evaluation of speaker attribution, diarization, role-aware performance (e.g., executives vs.\ analysts), and errors across industries and call structure. A comprehensive benchmark must therefore pair realistic long-form audio with rich metadata. 

\textbf{Earnings25} is designed to provide aligned transcripts together with speaker information, industry context, and structured metadata for fine-grained evaluation. Unlike prior resources that emphasize scale or aggregate metrics, it is explicitly constructed to reveal domain-specific and long-tail failure modes that are obscured by naturally distributed corpora. By combining long-form audio with industry-balanced evaluation, the benchmark enables systematic study of industry-aware ASR robustness and structured variation in financial speech.

\section{Prior Work}

\subsection{Conversational and meeting speech benchmarks}
Earnings calls share several conversational properties with meeting and telephone speech, including rapid turn-taking, disfluencies, and speaker overlap. Standard benchmarks such as \textbf{Switchboard} (1992)~\cite{godfrey1992switchboard}, \textbf{CallHome} (1997)~\cite{canavan1997callhome}, and \textbf{AMI} (2007)~\cite{carletta2007ami} capture aspects of these phenomena, but generally lack the finance-specific terminology, dense numeracy, and domain structure that significantly affect ASR performance on earnings calls.

\subsection{Existing financial speech benchmarks}
Existing financial speech datasets fall short as comprehensive evaluation benchmarks in several key dimensions. \textbf{SPGISpeech}, released in 2021 and expanded in 2025~\cite{spgispeech, spgispeech2}, provides over 5{,}000 hours of professionally transcribed earnings-call audio. However, it was designed primarily as a \emph{training corpus}: the dataset includes very large train, development, and test splits (with approximately 2{,}000 hours in the test set), making it computationally expensive for controlled and reproducible benchmark evaluation.

The \textbf{Earnings-21} (2021) and \textbf{Earnings-22} (2022) benchmarks~\cite{delrio21_interspeech,delrio2022earnings22} focus on open evaluation with aligned transcripts and accent- and country-level metadata. Earnings-22 provides approximately 160 hours of long-form global earnings calls, but lacks industry-balanced sampling: high-frequency sectors dominate evaluation metrics, obscuring performance on underrepresented domains. Moreover, neither dataset provides the structured metadata—such as speaker roles, industry classification, or company identifiers—needed for fine-grained error analysis and speaker-aware evaluation.

Overall, the financial speech domain lacks a comprehensive, balanced, and metadata-rich benchmark designed specifically for evaluating ASR systems under realistic earnings-call conditions.

\section{Corpus Definition}

Earnings25 comprises two complementary test sets designed to support both long-form and segment-level evaluation, as summarized in Table~\ref{tab:earnings25-testsets}. \textbf{testset-full} consists of 498 hours of complete earnings-call recordings from S\&P~500 companies in 2025~Q4, preserving full conversational context with typical call durations of approximately one hour. \textbf{testset-segmented} is a curated 46-hour evaluation set sampled from over 2{,}000 U.S. earnings calls across 2025 (Q1--Q4), comprising 290 segments—one per industry—to ensure balanced domain coverage. Segments in \textbf{testset-segmented} are 5--10 minutes long.

Both test sets include aligned transcripts and structured metadata and are constructed to preserve realistic conversational flow, natural turn-taking, and cross-speaker dynamics. In addition to transcription, Earnings25 provides rich metadata, including speaker segmentation, industry classification, and company identifiers, enabling fine-grained analysis beyond aggregate word error rate (WER) and supporting speaker- and structure-aware evaluation.

\begin{table*}[t]
  \caption{Overview of the Earnings25 test sets.}
  \label{tab:earnings25-testsets}
  \centering
  \begin{tabular}{lcc}
    \toprule
    \textbf{Property} & \textbf{testset-full} & \textbf{testset-segmented} \\
    \midrule
    Description & Full earnings calls & Industry-stratified segments \\
    Coverage & S\&P~500 companies & U.S. companies \\
    Date range & 2025~Q4 & January--December 2025 \\
    Total audio & 498 hours & 46 hours \\
    Units & \textasciitilde500 full calls & 290 segments (one per industry) \\
    Avg.~duration & 58 minutes & 9.5 minutes \\
    Countries & 12 (US: 93.8\%) & 1 (US only) \\
    Industries & 284 categories & 290 categories \\
    Sampling rate & 11--44.1 kHz (original) & 16 kHz \\
    Audio format & MP3 & WAV \\
    \bottomrule
  \end{tabular}
\end{table*}

\section{Corpus Generation}
We sample full earnings calls, apply CTC-based forced alignment to obtain word-level timestamps, and aggregate shorter speech segments into 5--10 minute evaluation units. This section details the industry-stratified sampling, alignment, and segmentation procedures.

\subsection{Industry-stratified sampling of the full earnings-call universe}
To construct a representative and industry-balanced evaluation set, we sample earnings calls from a pool of over 2{,}100 U.S. earnings calls spanning 2025~Q1--Q4. Sampling is performed using a two-stage, reproducible procedure designed to ensure broad industry coverage while preventing dominance by high-frequency sectors.

In the first stage, we filter the corpus to retain only U.S.-domiciled companies within the target date range. In the second stage, we apply \emph{disproportionate stratified sampling} based on industry classification. Calls are grouped by industry, and a fixed number of samples is drawn from each group, ensuring equal representation across industries regardless of their underlying frequency in the corpus. When the number of industry groups exceeds the target evaluation size, a final random selection is applied. All sampling steps are seeded for reproducibility.

\subsection{Alignment}
We perform forced alignment to obtain word-level timestamps, which enable the extraction of shorter segments from long earnings calls and thereby increase industry coverage. These time spans are also used to support speaker diarization and speaker-aware analysis.

\subsubsection{CTC-based forced alignment}
Forced alignment is performed using Connectionist Temporal Classification (CTC) models implemented in NVIDIA NeMo~\cite{kuchaiev2019nemo}. CTC~\cite{graves2006ctc} defines a sequence-level objective that marginalizes over all monotonic alignments between input acoustic frames and output label sequences, producing frame-level posterior probabilities over output tokens and a blank symbol. These posteriors are used to align reference transcripts to audio and recover word-level timing information.

Given a reference transcript tokenized into subword units, we compute frame-level CTC log-probabilities and apply Viterbi decoding to recover the most likely alignment path under the CTC lattice. Token-level timestamps are then aggregated into word boundaries. To mitigate alignment errors caused by trailing silence or low-confidence regions, we impose a maximum word-duration constraint of 0.5 seconds.

The alignment pipeline consists of: (1) resampling audio to 16\,kHz WAV format; (2) computing frame-level CTC log-probabilities; (3) Viterbi alignment between audio frames and reference tokens; and (4) word-boundary extraction from aligned tokens.

\subsection{Segment extraction from full calls}
After obtaining word-level timestamps via forced alignment, we aggregate contiguous speech segments into non-overlapping \emph{blocks} with durations constrained to a predefined range (e.g., 5--10 minutes). Blocks are constructed greedily by accumulating consecutive segments until a minimum duration is reached and extending the block while the total duration remains below the maximum threshold.

\subsubsection{Quality filtering}
We apply content-based filtering to remove blocks whose merged transcripts contain undesired patterns such as operator boilerplate or non-speech cues. From the remaining candidates, one block is sampled uniformly at random per call to avoid over-representation of individual calls. Prior to audio extraction, a fixed padding of 0.2 seconds is added to block boundaries to mitigate alignment jitter. This procedure produces a compact and diverse set of segments with clean transcripts, balanced across calls and suitable for segment-level ASR benchmarking.

Speaker tags are used to further split the segments to ensure that segment boundaries do not bisect speaker turns. This procedure yields transcripts with reliable timing suitable for segment-level ASR evaluation, speaker-aware analysis, and downstream speech processing tasks.

\section{Corpus Analysis}

\subsection{Geographic coverage and accent distribution}
\textbf{testset-full} includes earnings calls from all S\&P~500 companies and spans a broad geographic footprint, covering English earnings calls from 12 countries. While U.S.-domiciled companies dominate the corpus (reflecting index composition), the dataset also includes calls from companies headquartered in Europe, Asia, and other regions. This geographic diversity introduces English accent variation and heterogeneous recording conditions that reflect real-world earnings-call audio.

In contrast, \textbf{testset-segmented} restricts evaluation to U.S.-domiciled companies to reduce accent variability and provide a controlled evaluation setting. This design allows baseline ASR performance to be assessed under consistent linguistic conditions, while accent-robustness can be studied using the full-call test set.

\begin{table}[t]
  \caption{Geographic distribution of \textbf{testset-full}.}
  \label{tab:geo-dist}
  \centering
  \begin{tabular}{lc}
    \toprule
    \textbf{Country} & \textbf{Number of calls} \\
    \midrule
    United States & 482 \\
    Ireland & 8 \\
    United Kingdom & 6 \\
    India & 5 \\
    Switzerland & 4 \\
    Australia & 2 \\
    Bermuda & 2 \\
    Netherlands & 1 \\
    Mexico & 1 \\
    Mauritius & 1 \\
    China & 1 \\
    Canada & 1 \\
    \bottomrule
  \end{tabular}
\end{table}

\subsection{Industry distribution}
Industry coverage is a core design consideration of \textbf{Earnings25}. \textbf{testset-full} spans 284 distinct industry categories, reflecting the natural sector distribution of the S\&P~500. As summarized in Table~\ref{tab:industry-full}, utilities and financial services are among the most frequently represented sectors, while the majority of industries appear only a small number of times, preserving the long-tail structure characteristic of financial data.

\begin{table}[t]
  \caption{Industry distribution of \textbf{testset-full}.}
  \label{tab:industry-full}
  \centering
  \begin{tabular}{lc}
    \toprule
    \textbf{Industry category} & \textbf{Number of calls} \\
    \midrule
    Integrated Electric Utilities & 21 \\
    Financial Transaction Processors & 9 \\
    Biotech & 9 \\
    Banks & 9 \\
    Enterprise Software & 8 \\
    Crude Oil \& Natural Gas E\&P & 7 \\
    Apartment REIT & 7 \\
    Large Pharma & 6 \\
    Infrastructure Software & 6 \\
    Diversified Industrials & 6 \\
    Security \& Cmdty Exchanges & 5 \\
    P\&C Insurance Premiums & 5 \\
    Life Insurance & 5 \\
    Investment Management & 5 \\
    Other (270 categories) & 406 \\
    \bottomrule
  \end{tabular}
\end{table}

To mitigate frequency bias in evaluation, \textbf{testset-segmented} is constructed via industry-stratified sampling from over 2{,}000 U.S. earnings calls spanning 2025 (Q1--Q4). All 290 segments are drawn exclusively from U.S.-domiciled companies to ensure consistent audio quality and reduce accent variability. The segmented test set spans 290 unique industry categories, with exactly one segment per industry, ensuring broad coverage of domain-specific vocabulary across diverse financial sub-domains—from 3D Printers and Adult Nightclubs to Wind Turbines and Wireline Telecom Equipment. This design enables fine-grained, industry-aware analysis of ASR performance while maintaining a controlled linguistic evaluation setting.

\subsection{Speaking-style variation}
Earnings calls exhibit substantial variation in speaking style and interaction structure. Within a single call, speech alternates between scripted prepared remarks and spontaneous analyst Q\&A, often with rapid turn-taking and occasional overlap. \textbf{testset-full} preserves these dynamics across entire calls, while \textbf{testset-segmented} retains multi-speaker conversational structure within each 5--10 minute segment.

Speaker attribution metadata enables role-aware analysis of ASR performance across operators, executives, and analysts. Operators typically deliver formulaic announcements, executives present prepared remarks, and analysts contribute unscripted questions. This variation allows evaluation of ASR robustness across speaking styles, roles, and interaction patterns common in financial speech.


\begin{table*}[t]
  \caption{Baseline ASR results on the full earnings-call test set (498h) and the curated segmented test set (46h). Lower is better.}
  \label{tab:bench}
  \centering
  \begin{tabular}{lcccccccc}
    \toprule
     & \multicolumn{4}{c}{\textbf{testset-full}} & \multicolumn{4}{c}{\textbf{testset-segmented}} \\
    \cmidrule(lr){2-5} \cmidrule(lr){6-9}
    \textbf{ASR Model} 
    & \textbf{WER} 
    & \textbf{WER-N} 
    & \textbf{WER-nc-np} 
    & \textbf{WER-N-nc-np}
    & \textbf{WER} 
    & \textbf{WER-N} 
    & \textbf{WER-nc-np} 
    & \textbf{WER-N-nc-np} \\
    \midrule
    Whisper-base 
      & 0.17853 & 0.17068 & 0.11572 & 0.11043 
      & 0.19373 & 0.18604 & 0.1202 & 0.11606 \\
    Whisper-medium 
      & 0.14399 & 0.13659 & 0.08594 & 0.08151 
      & 0.15536 & 0.14932 & 0.08676 & 0.08369 \\
    Whisper-large-v2 
      & 0.14039 & 0.13407 & 0.08422 & 0.08036 
      & 0.15333 & 
0.14724 & 0.08459 & 0.08174 \\
    Parakeet-tdt-0.6b-v2 
      & 0.10837 & 0.10326 & 0.06413 & \textbf{0.06114} 
      & 0.11109 & 0.10498 & 0.06473 & \textbf{0.06062} \\
    \bottomrule
  \end{tabular}
\end{table*}

\begin{table*}[t]
\centering
\caption{Selected industry-level ASR performance of \textbf{Parakeet-tdt-0.6b-v2} on \textbf{testset-full} (2--3 calls per subsector based on industry tags). This table is illustrative and not a comprehensive ranking. Lower is better.}
\label{tab:industry_parakeet_sorted}
\begin{tabular}{lcccc}
\toprule
\textbf{Industry} & \textbf{WER} & \textbf{WER-N} & \textbf{WER-nc-np} & \textbf{WER-N-nc-np} \\
\midrule
Biotech & 0.15440 & 0.14710 & 0.10265 & 0.09845 \\
Pharma & 0.15343 & 0.14821 & 0.10255 & 0.10097 \\
Crude Oil \& Natural Gas E\&P & 0.13131 & 0.12576 & 0.07759 & 0.07416 \\
Health Care Software & 0.13033 & 0.12347 & 0.07849 & 0.07466 \\
\bottomrule
\end{tabular}
\end{table*}

\section{Transcription Experiments}

\subsection{Baseline Models and Evaluation Protocol}

We benchmark representative ASR systems spanning sequence-to-sequence and transducer architectures: OpenAI \textbf{Whisper} models~\cite{radford2022whisper} (base, medium, large-v2) and NVIDIA NeMo’s \textbf{Parakeet-TDT-0.6B-v2}~\cite{xu2023efficient}. No external language model is used.\\
\textbf{Whisper inference:}
Decoding options are loaded from a fixed model configuration. We set a fixed random seed and use deterministic decoding settings. The language is set to English. No optional keyword prompting or boosting words are used during decoding.\\
\textbf{Parakeet-TDT inference:}
We use the pretrained checkpoint \texttt{nvidia/parakeet-tdt-0.6b-v2} with greedy transducer decoding and no external language model.\\
\textbf{Scoring and normalization:}
We report four consistent variants:
\textbf{WER} (raw), \textbf{WER-N} (NeMo-normalized), \textbf{WER-nc-np} (lowercased, punctuation removed), and \textbf{WER-N-nc-np} (NeMo-normalized, lowercased, punctuation removed).
For normalization-based variants, the same NeMo English text normalization~\cite{zhang2021nemo} is applied to both reference and hypothesis.\\
\textbf{Reproducibility:}
All sampling and segmentation procedures use a fixed random seed (2025). Experimental scripts set \texttt{PYTHONHASHSEED=2025} and framework RNG seeds, and log the full decoding configuration used for each run.

\subsection{Results}

Table~\ref{tab:bench} reports baseline performance on \textbf{testset-full} (498 h) and the industry-balanced \textbf{testset-segmented} (46 h). Lower is better.
Across models, performance on \textbf{testset-segmented} is slightly worse than on \textbf{testset-full}, despite shorter duration. This reflects the effect of industry stratification: while \textbf{testset-full} follows the natural frequency distribution of sectors, the segmented set enforces equal industry representation and increases exposure to long-tail terminology.

To illustrate domain variability, Table~\ref{tab:industry_parakeet_sorted} reports results for selected subsectors (2--3 calls each) using Parakeet-tdt-0.6b-v2. This is an illustrative subset and not a comprehensive industry ranking. Terminology-dense domains such as biotech and pharma show substantially higher WER (15.3--15.4\%) than the aggregate testset-full WER (10.8\%).
Because the corpus-level metric is frequency-weighted across industries, high-volume and more repetitive sectors lower the overall average. In contrast, the unweighted per-industry results emphasize more challenging domains, demonstrating that aggregate WER can mask substantial domain-dependent variation.

\section{Limitations and Conclusion}
Earnings25 provides value along three dimensions: (1) \textbf{domain-specific evaluation}, offering a challenging benchmark based on S\&P~500 earnings calls with broad industry coverage and finance-specific terminology; (2) \textbf{reproducible baselines}, with standardized evaluations for contemporary ASR models, including Whisper and Parakeet-TDT; and (3) \textbf{rich metadata}, including industry, call-structure, and speaker annotations that enable stratified and speaker-aware analysis.

Regarding limitations, Earnings25 focuses on English-language earnings calls and primarily reflects speech from U.S.-domiciled companies. While this design enables controlled and high-quality evaluation, it does not capture the full linguistic diversity of global earnings calls. Extending the benchmark to include multilingual earnings-call data across a broader range of countries and languages remains an important direction for future work.






\section{Data Access and Licensing}

Earnings25 is released for research and benchmarking purposes. We redistribute the audio recordings, transcripts, metadata, annotations, and evaluation splits through Zenodo:
\href{https://doi.org/10.5281/zenodo.18762168}
{\textcolor{blue}{https://doi.org/10.5281/zenodo.18762168}}.
The transcripts, annotations, metadata, evaluation splits, and alignments are released under the Creative Commons Attribution 4.0 International license (CC BY 4.0). The redistributed audio recordings remain subject to any applicable terms of the original content providers, and users are responsible for ensuring compliance with those terms.

\section{Generative AI Use Disclosure}
During preparation of this manuscript, the authors used a generative AI assistant only for language editing and polishing (e.g., grammar and phrasing). The tool was not used to generate experimental results, analyses, or conclusions, and no generative AI system is listed as an author.

\bibliographystyle{IEEEtran}
\bibliography{mybib}

\end{document}